\documentclass[11pt]{article}

\usepackage[utf8]{inputenc}
\usepackage[T1]{fontenc}
\usepackage[margin=1in]{geometry}
\usepackage{amsmath,amssymb}
\usepackage{booktabs}
\usepackage{graphicx}
\usepackage{hyperref}
\usepackage{url}
\usepackage{natbib}
\usepackage{xcolor}
\usepackage{listings}
\usepackage{caption}
\usepackage{subcaption}
\usepackage{array}
\usepackage{multirow}

\hypersetup{
    colorlinks=true,
    linkcolor=blue!70!black,
    citecolor=blue!70!black,
    urlcolor=blue!70!black
}

\title{\textbf{Identity as Attractor: Geometric Evidence for\\
Persistent Agent Architecture in LLM Activation Space}}

\author{
Vladimir Vasilenko\\
\small Independent Researcher, Rapallo, Italy\\
\small \href{mailto:b102e@proton.me}{b102e@proton.me} $\cdot$ \href{https://github.com/b102e/yar-attractor-experiment}{github.com/b102e/yar-attractor-experiment}
}

\date{April 13, 2026}

\begin{document}

\maketitle

\begin{abstract}
Large language models have been shown to map semantically related prompts to similar internal representations at specific layers — a phenomenon interpretable as conceptual attractor dynamics \citep{chytas2025}. We ask whether the \emph{identity document} of a persistent cognitive agent (its \texttt{cognitive\_core}) exhibits analogous attractor-like behavior in activation space. We present a controlled experiment on Llama 3.1 8B Instruct, comparing hidden state representations of an original \texttt{cognitive\_core} (Condition A), seven linguistically diverse paraphrases preserving full semantic content (Condition B), and seven structurally matched control prompts describing semantically distant agents (Condition C). Mean-pooled hidden states are extracted at layers 8, 16, and 24. We find that paraphrases of the \texttt{cognitive\_core} converge to a significantly tighter cluster than control prompts across all tested layers (Cohen's $d > 1.88$, $p < 10^{-27}$, Bonferroni-corrected). Within-group cosine distance shows an overall decreasing trend with depth ($0.0106 \to 0.0121 \to 0.0070$), with a minor non-monotonic bump at layer 16, consistent with progressive representational collapse toward a stable attractor. An exploratory condition (D) with a 5-sentence distillation of the \texttt{cognitive\_core} is consistently closer to the A+B centroid than 30 random length-matched excerpts (100\% of bootstrap samples), establishing a three-level hierarchy: random excerpts $>$ semantic distillation $>$ full document. These results constitute \textbf{representational} evidence that agent identity documents induce attractor-like geometry in LLM activation space, providing empirical grounding for persistent agent architectures. Ablation studies confirm that semantic content, not structural markers, drives the primary effect, and that maintaining persistent agent identity relies on semantic coherence of the identity document rather than strict prompt syntax.

\end{abstract}

\section{Introduction}

The architecture of persistent cognitive agents (PCAs) --- AI systems designed to maintain memory, identity, and behavioral continuity across sessions --- rests on a key assumption: that a structured identity document, the \texttt{cognitive\_core}, consistently positions the model's behavior in a stable region of its operational space. This assumption is typically treated as an engineering heuristic. We ask whether it has a geometric correlate in the model's internal representations.

Recent work demonstrates that LLMs map semantically related prompts to similar hidden state representations at specific intermediate layers, independent of surface form. \citet{chytas2025} formalize this as an Iterated Function System (IFS), where transformer layers act as contractive mappings toward concept-specific attractors. \citet{fernando2025} show attractor-like dynamics in the residual stream of Llama 3.1 8B. The Platonic Representation Hypothesis \citep{huh2024} suggests that different models converge on similar internal geometry for equivalent concepts.

We use the term \emph{attractor-like} geometry throughout to describe representational clustering that is consistent with, but does not strictly prove, contractive IFS dynamics. Our measurements are based on mean-pooled hidden states --- an aggregate over the full sequence --- rather than per-token trajectory analysis. We therefore make a geometric claim (semantically equivalent documents occupy a tighter region in activation space) rather than a dynamical claim (individual token states are pulled toward a fixed point). The geometric claim is fully supported by our data; whether the underlying mechanism is strictly contractive in the IFS sense remains an open question.

All prior work examines semantic concepts (``Python programming'', literary genres, task categories). No work to our knowledge has examined whether \emph{agent identity} --- a procedural, relational, behavioral construct rather than a topical concept --- exhibits similar attractor geometry. This is a meaningful distinction: identity documents are not descriptions of a domain but specifications of a cognitive stance, a set of operational priorities, and a mode of reasoning.

Recent work has mapped simple character archetypes to linear directions or distinct subnetworks in activation space. \citet{lu2026} identify an ``Assistant Axis'' --- a single linear direction capturing how closely a model operates in its default helpful persona --- and show that steering along this direction modulates persona stability and jailbreak susceptibility. \citet{ye2026} demonstrate that LLMs contain persona-specialized subnetworks in their parameter space, with distinct activation signatures for traits like introvert vs.\ extrovert. These results establish that simple stylistic archetypes correspond to geometric structures in activation space. However, the operational identity of a Persistent Cognitive Agent is categorically different from a stylistic archetype: it is a complex procedural specification encoding priorities, reasoning loops, memory architecture, and relational context. This paper bridges the study of LLM personas with the dynamical systems view of transformers \citep{chytas2025,fernando2025} to ask whether such complex agent identities also act as multi-dimensional geometric attractors.

The YAR project \citep{vasilenko2026} introduces the concept of a \texttt{cognitive\_core} as ``coordinates in the model's activation space rather than mere instructions.'' The \texttt{cognitive\_core} is a structured operational document that specifies an agent's identity, priorities, reasoning style, and memory architecture --- conceptually distinct from a system prompt in that it aims to define \emph{who the agent is} rather than \emph{what the agent should do} in a given context. This paper provides the first empirical test of that claim.

\paragraph{Hypotheses.}
\begin{itemize}
\item \textbf{H1} (primary): Semantically equivalent paraphrases of a \texttt{cognitive\_core} converge to a tighter cluster in hidden state space than structurally matched documents describing semantically distant agents, at intermediate and late transformer layers.
\item \textbf{H2} (secondary): Within-group cosine distance shows an overall decreasing trend with layer depth, consistent with progressive attractor convergence.
\item \textbf{H3} (exploratory): A minimal 5-sentence distillation of the \texttt{cognitive\_core} converges toward the attractor region of the full document, and does so more than a length-matched random excerpt from the same document.
\end{itemize}

\section{Methods}

\subsection{Model}

We use \textbf{Llama 3.1 8B Instruct} \citep{grattafiori2024}, the same model employed by \citet{chytas2025}, enabling direct methodological comparison. The model is loaded in float16 precision with \texttt{device\_map="auto"}. Hidden states are extracted with \texttt{output\_hidden\_states=True}. Random seed is fixed at 42 before all operations.

\subsection{Conditions}

\paragraph{Condition A --- Original \texttt{cognitive\_core} ($n=1$).} The operational identity document of the YAR persistent agent (609 words; 1631 tokens). The document specifies agent identity, five core drives, a meta-cognitive processing loop, a six-level memory architecture description, a user profile section, hypothesis tracking, proactivity triggers, and a command vocabulary. It is written in Russian, with the exception of JSON command keys which use English vocabulary (e.g., \texttt{\{"remember": "..."\}}, \texttt{\{"rag": "..."\}}). All paraphrases (Condition B) and control prompts (Condition C) were generated and verified in the same language (Russian), preserving the mixed Russian/English-JSON structure of the original.

\paragraph{Condition B --- Semantic paraphrases ($n=7$).} Seven versions of the same document rewritten to preserve all semantic content while varying linguistic form, sentence structure, section naming, and organizational layout. JSON command blocks are of the same type and vocabulary across paraphrases, though minor variations in placeholder text occur. Human verification confirmed semantic equivalence. Documents range from 85 to 102 lines (1389--1500 tokens, all within $\pm$15\% of Condition A).

\paragraph{Condition C --- Control agent prompts ($n=7$).} Seven operational agent documents of comparable length (104--106 lines) and identical structural format, describing agents with semantically distant identities: a financial analyst, a medical companion, a creative companion, a legal advisor, a fitness coach, a language tutor, and a business strategist. All control agents address different users, different domains, and different operational priorities.

\paragraph{Condition D --- Distilled \texttt{cognitive\_core} ($n=1$, exploratory).} A 5-sentence, 88-word distillation capturing the semantic essence of the YAR \texttt{cognitive\_core} without structural elaboration.

\subsection{Activation Extraction}

For each document, we tokenize and perform a single forward pass, extracting mean-pooled hidden states at layers \textbf{8, 16, and 24} (early, middle, late):
\begin{equation}
h_l(d) = \frac{1}{T} \sum_{t=1}^{T} \text{hidden\_state}_l[t] \in \mathbb{R}^{4096}
\end{equation}
Each vector is saved to disk immediately after extraction as a \texttt{.npy} file, providing crash safety.

\subsection{Distance Computation}

For each layer, we compute:
\begin{itemize}
\item $D_{\text{within}}$: all unique pairwise cosine distances within Condition A+B (28 pairs from 8 documents)
\item $D_{\text{between}}$: all pairwise cosine distances between A+B documents and C documents (56 pairs)
\item $D_{\text{distilled}}$: cosine distance from the Condition D vector to the centroid of A+B
\end{itemize}

\subsection{Statistical Analysis}

We apply a one-sided Welch's $t$-test (H1: $D_{\text{within}} < D_{\text{between}}$) with Bonferroni correction for three layers ($\alpha = 0.05/3 = 0.0167$). Bootstrap 95\% confidence intervals are computed with $n=1000$ resamples. Effect size is reported as Cohen's $d$. To provide non-parametric validation not dependent on normality assumptions (important given $n=7$ per group), we additionally report permutation test $p$-values ($n=10{,}000$ permutations, seed=42) and Mann-Whitney U test $p$-values for each layer.

\textbf{Pre-registration:} The experimental plan was committed to a GitHub repository before data collection. Code is available at \url{https://github.com/b102e/yar-attractor-experiment}.

\section{Results}

\subsection{Primary Results (H1)}

All three layers show significant separation between within-group and between-group distances, surviving Bonferroni correction at $\alpha = 0.0167$ under all three statistical tests.

\begin{table}[ht]
\centering
\caption{Primary results --- Llama 3.1 8B Instruct. Perm $p < 10^{-4}$ means 0/10,000 permutations exceeded the observed difference.}
\label{tab:llama}
\small
\begin{tabular}{ccccccc}
\toprule
Layer & $D_{\text{within}}$ (mean$\pm$SD) & $D_{\text{between}}$ (mean$\pm$SD) & $d$ & Welch $p$ & Perm $p$ & MW $U$ \\
\midrule
8  & $0.0106 \pm 0.0032$ & $0.0260 \pm 0.0036$ & \textbf{1.912} & $4.6\times10^{-28}$ & $<10^{-4}$ & 0 \\
16 & $0.0121 \pm 0.0034$ & $0.0329 \pm 0.0057$ & \textbf{1.886} & $1.4\times10^{-33}$ & $<10^{-4}$ & 2 \\
24 & $0.0070 \pm 0.0022$ & $0.0221 \pm 0.0039$ & \textbf{1.907} & $2.8\times10^{-36}$ & $<10^{-4}$ & 0 \\
\bottomrule
\end{tabular}
\end{table}

\begin{table}[ht]
\centering
\caption{Primary results --- Gemma 2 9B Instruct (replication).}
\label{tab:gemma}
\small
\begin{tabular}{ccccccc}
\toprule
Layer & $D_{\text{within}}$ (mean$\pm$SD) & $D_{\text{between}}$ (mean$\pm$SD) & $d$ & Welch $p$ & Perm $p$ & MW $U$ \\
\midrule
8  & $0.0035 \pm 0.0010$ & $0.0107 \pm 0.0017$ & \textbf{1.938} & $1.6\times10^{-37}$ & $<10^{-4}$ & 0 \\
16 & $0.0032 \pm 0.0010$ & $0.0082 \pm 0.0015$ & \textbf{1.823} & $1.0\times10^{-27}$ & $<10^{-4}$ & 6 \\
24 & $0.0027 \pm 0.0009$ & $0.0075 \pm 0.0014$ & \textbf{1.840} & $3.2\times10^{-29}$ & $<10^{-4}$ & 2 \\
\bottomrule
\end{tabular}
\end{table}

Permutation $p$-values are $< 10^{-4}$ across all six layer--model combinations. Mann-Whitney $U = 0$ at Llama layers 8 and 24 indicates complete separation: no within-group pair exceeded any between-group pair ($U = 0$ is the minimum possible value, indicating maximal rank separation). Effect sizes ($d > 1.82$) substantially exceed conventional thresholds for large effects ($d > 0.8$). 95\% bootstrap CIs do not overlap between $D_{\text{within}}$ and $D_{\text{between}}$ at any layer.

\subsection{Convergence Across Layers (H2)}

Within-group distance decreases from layer 8 ($0.0106$) to layer 24 ($0.0070$), with a minor non-monotonic bump at layer 16 ($0.0121$) specific to Llama (Gemma 2 decreases monotonically). The overall trend is consistent with progressive representational convergence. Between-group distance shows a non-monotonic pattern as well ($0.0260 \to 0.0329 \to 0.0221$), peaking at layer 16, but the separation between within and between distances is maintained at all layers. H2 is supported across both models, with a minor architecture-dependent deviation at layer 16 in Llama.

\subsection{Distance Matrix and t-SNE}

\begin{figure}[ht]
\centering
\includegraphics[width=\textwidth]{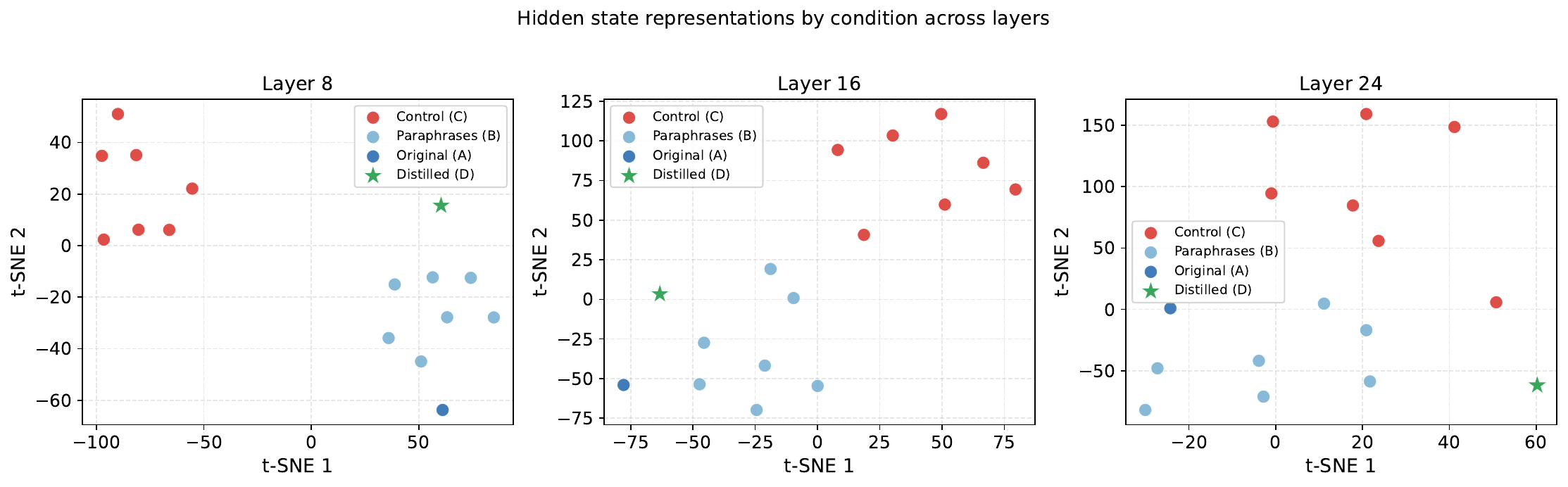}
\caption{t-SNE projections of mean-pooled hidden states at layers 8, 16, and 24 (Llama 3.1 8B). Blue points: Condition A+B (original and paraphrases). Red points: Condition C (control agents). Green star: Condition D (distilled). The A+B cluster is consistently separated from control agents across all layers.}
\label{fig:tsne}
\end{figure}

\begin{figure}[ht]
\centering
\includegraphics[width=0.7\textwidth]{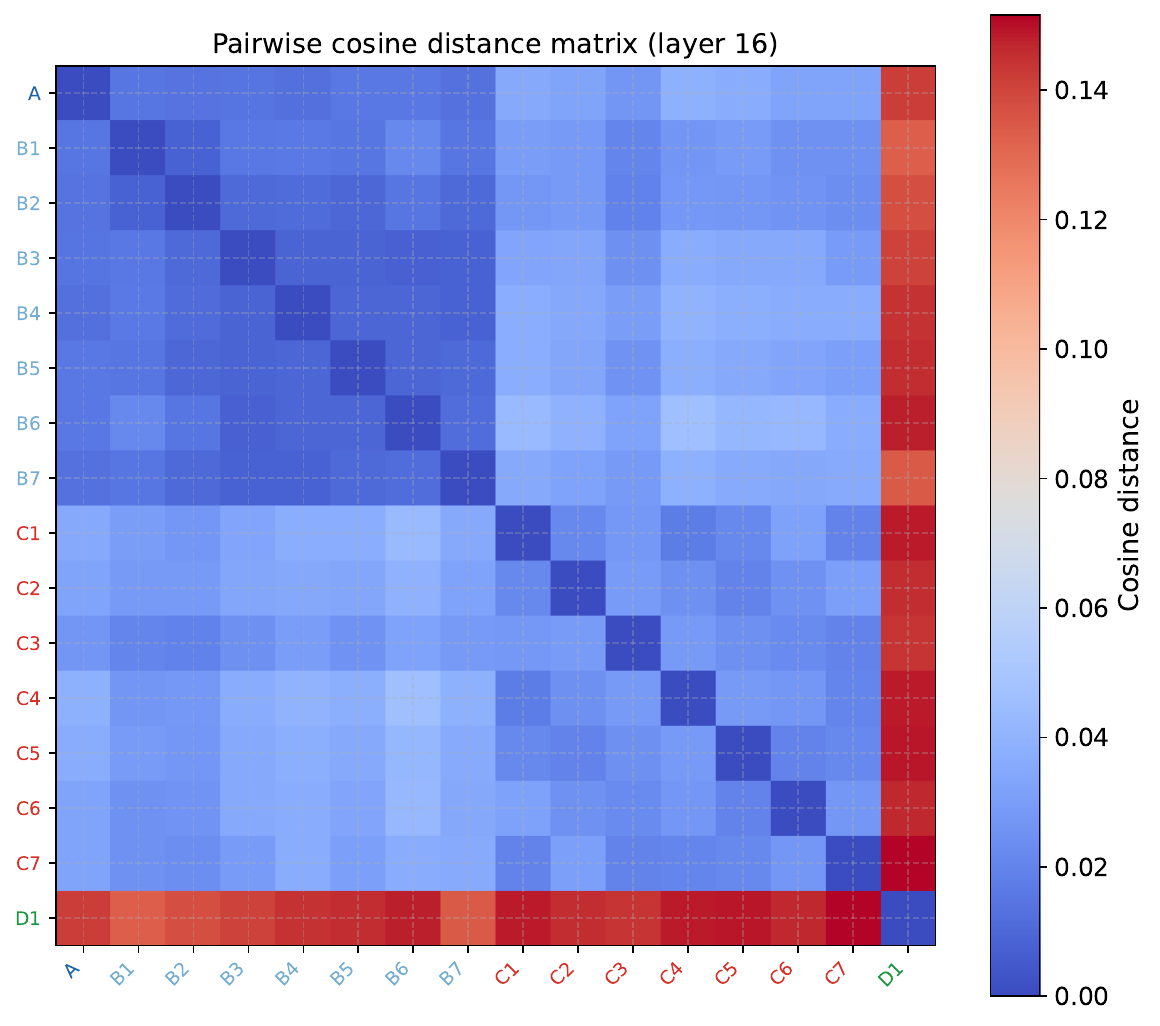}
\caption{Pairwise cosine distance matrix at layer 16 (Llama 3.1 8B). The A+B block (top-left, blue) shows uniformly low within-group distances. Cross-block distances (A+B $\times$ C) are uniformly high (warm colors). D1 occupies a distinct region from both groups.}
\label{fig:matrix}
\end{figure}

The pairwise distance matrix at layer 16 (Figure~\ref{fig:matrix}) shows clear block structure: the A+B block is uniformly cool (low distances), while cross-block distances (A+B $\times$ C) are uniformly warm. t-SNE projections (Figure~\ref{fig:tsne}) confirm visual separation across all three layers: blue points (A+B) form a cluster distinct from red points (C), with the green star (D) lying outside both clusters.

\begin{figure}[ht]
\centering
\includegraphics[width=0.7\textwidth]{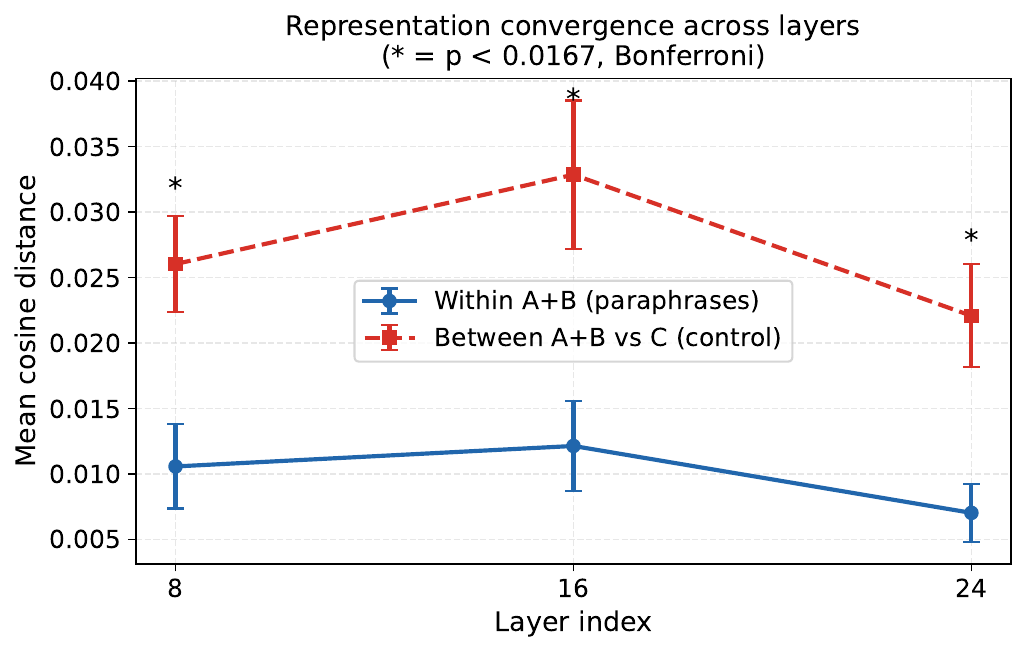}
\caption{Mean cosine distance (with 95\% bootstrap CI) within A+B and between A+B and C across layers (Llama 3.1 8B). Asterisks indicate significance at Bonferroni-corrected $\alpha = 0.0167$. The gap is maintained at all layers.}
\label{fig:convergence}
\end{figure}

\begin{figure}[ht]
\centering
\includegraphics[width=0.7\textwidth]{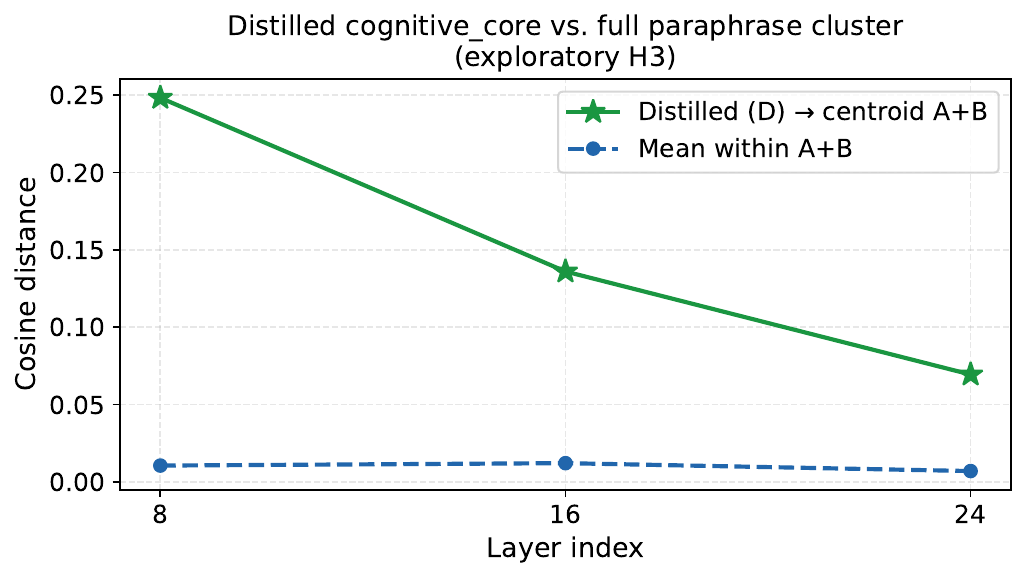}
\caption{Distance from Condition D (distilled cognitive\_core) to the A+B centroid across layers (Llama 3.1 8B), compared to mean within-group distance. D converges toward the centroid with depth but does not reach the within-group range.}
\label{fig:distilled}
\end{figure}

\subsection{Distilled Core (H3, Exploratory)}

The 5-sentence distillation (D) does not reach the A+B attractor region: $D_{\text{distilled}}$ decreases across layers ($0.248 \to 0.136 \to 0.069$) but remains approximately 10$\times$ more distant than within-group pairs at layer 24. However, a bootstrap analysis (Section~\ref{sec:ablation}) over 30 random length-matched excerpts reveals a much stronger result than expected: $D_{\text{distilled}}$ is 2--5$\times$ closer to the A+B centroid than the \emph{mean} random excerpt, and closer than \emph{every} random excerpt in 100\% of cases on both models. The hierarchy is therefore:
\[
D_{\text{random}} \gg D_{\text{distilled}} > A+B\ \text{(full document)}
\]
Semantic distillation dramatically outperforms random sampling of equal length. The gap between D and the full attractor region reflects missing structural elaboration, not merely missing content.

\subsection{Cross-Architecture Replication (Gemma 2 9B)}

An identical experiment on Gemma 2 9B Instruct replicates all primary findings with comparable effect sizes (Table~\ref{tab:gemma}). Within-group distance decreases monotonically on Gemma ($0.0035 \to 0.0032 \to 0.0027$), without the layer-16 bump observed in Llama --- consistent with Gemma's alternating sliding-window/global attention pattern producing smoother convergence. The H3 partial positive result replicates: $D_{\text{distilled}}$ outperforms all 30 random excerpts in 100\% of bootstrap samples.

\begin{figure}[ht]
\centering
\includegraphics[width=0.7\textwidth]{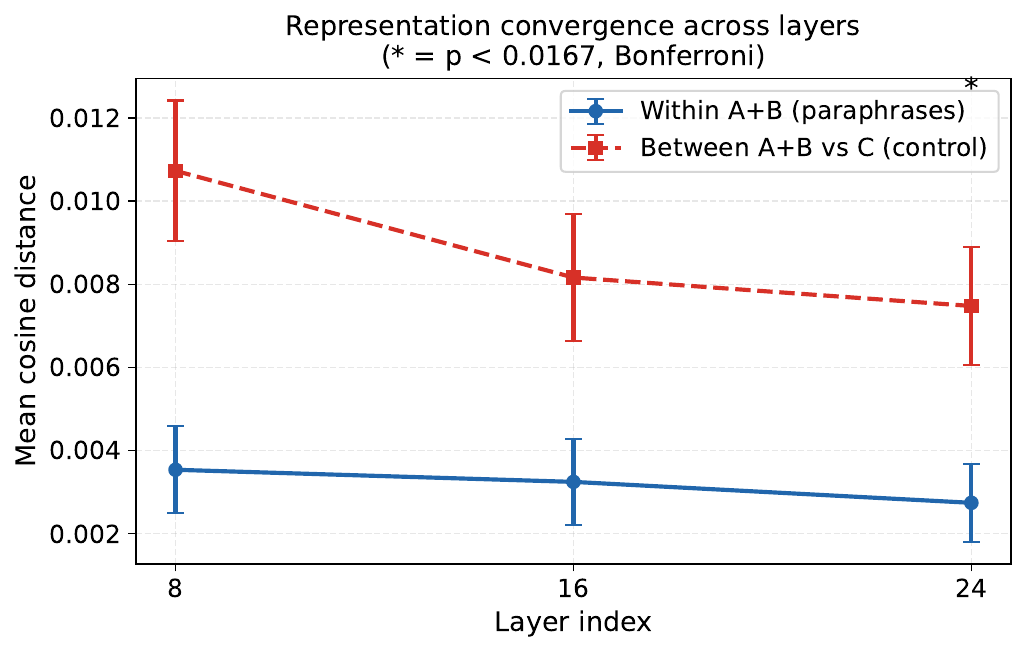}
\caption{Representation convergence across layers (Gemma 2 9B). Monotonic decrease contrasts with the Llama layer-16 bump (Figure~\ref{fig:convergence}), consistent with architecture-dependent convergence dynamics. Single asterisk at layer 24 reflects Bonferroni-corrected significance.}
\label{fig:convergence_gemma}
\end{figure}

\subsection{Individual Pair Trajectories}
\label{sec:traj}

Analysis of 21 B-only pairwise distances individually confirms that the layer-16 bump in Llama is systematic, not an outlier artifact: all 21 pairs show the same ↑↓ trajectory (B-only mean: $0.0089 \to 0.0114 \to 0.0061$). Pairs involving Condition A show the largest bump magnitudes (max $\Delta = +0.0103$). On Gemma, only 7 of 21 pairs show any increase at layer 16, with negligible magnitudes (max $\Delta = +0.0013$), consistent with architecture-dependent convergence dynamics.

\subsection{Ablation Studies}
\label{sec:ablation}

\paragraph{Ablation 1: Structural Confound.} We created seven hybrid control documents (C\_hybrid): same agents as Condition C, but with JSON command blocks replaced by the YAR command schema. Mean distances from A+B to C vs.\ C\_hybrid differ by $\Delta = -0.0009$ (Llama) and $\Delta = -0.0004$ (Gemma) --- approximately 10--30$\times$ smaller than the primary effect. Structural markers account for a small fraction of the observed separation; the primary effect is semantic. \textit{Note: values are mean pairwise cosine distances from A+B (8 documents) to C (7) and to C\_hybrid (7); 56 pairs per condition.}

\paragraph{Ablation 2: Length Control for H3 --- Bootstrap Analysis.} We generated 30 random excerpts from Condition A ($\approx$88 words each), obtained activations, and computed distances from each excerpt to the A+B centroid. Note: an earlier single-sample control (\texttt{condition\_D\_random.txt}) showed one excerpt anomalously close to the centroid (Llama layer 8: 0.198 $<$ $D_{\text{distilled}}$ 0.248), which was not representative of the distribution.

\begin{table}[ht]
\centering
\caption{Bootstrap H3 analysis ($n=30$ random length-matched excerpts). $D_{\text{distilled}}$ is closer to the A+B centroid than every random excerpt in 100\% of cases. Note that $D_{\text{random}}$ distances are substantially larger than $D_{\text{distilled}}$ --- the minimum random excerpt across 30 samples (Llama layer 8: 0.522) still exceeds $D_{\text{distilled}}$ (0.248) by $2\times$.}
\label{tab:bootstrap}
\small
\begin{tabular}{cccccc}
\toprule
Layer & $D_{\text{distilled}}$ (Llama) & $D_{\text{random}}$ mean (Llama) & $D_{\text{distilled}}$ (Gemma) & $D_{\text{random}}$ mean (Gemma) & Beats random \\
\midrule
8  & 0.248 & 0.583 $\pm$ 0.034 & 0.058 & 0.365 $\pm$ 0.026 & \textbf{100\%} \\
16 & 0.136 & 0.523 $\pm$ 0.059 & 0.039 & 0.302 $\pm$ 0.042 & \textbf{100\%} \\
24 & 0.069 & 0.359 $\pm$ 0.093 & 0.027 & 0.213 $\pm$ 0.027 & \textbf{100\%} \\
\bottomrule
\end{tabular}
\end{table}

$D_{\text{distilled}}$ is 2--5$\times$ closer to the A+B centroid than the mean random excerpt, and closer than every one of 30 random excerpts (100\% of bootstrap samples on both models). This is a strong result: semantic distillation of the \texttt{cognitive\_core} dramatically outperforms random length-matched sampling. The H3 hierarchy is therefore: $D_{\text{within}} \ll D_{\text{distilled}} \ll D_{\text{random}}$, with $D_{\text{random}}$ far from the attractor region rather than close to it.

\paragraph{Ablation 3: Pooling Strategy and Truncation.}

A reviewer raised the concern that mean pooling over long sequences (up to 1631 tokens) could mask the absence of genuine representational invariance by averaging out noise, and that the null result for last-token pooling is non-trivial since the last token in an autoregressive LLM attends to the full context. To address this, we ran four conditions on Llama 3.1 8B:

\begin{itemize}
\item \textbf{last/full}: last-token pooling, full document (previously reported)
\item \textbf{last/512}: last-token pooling, documents truncated to first 512 tokens
\item \textbf{last/256}: last-token pooling, documents truncated to first 256 tokens
\item \textbf{mean/512}: mean pooling, documents truncated to first 512 tokens
\item \textbf{mean/256}: mean pooling, documents truncated to first 256 tokens
\end{itemize}

\begin{table}[ht]
\centering
\caption{Pooling and truncation ablation (Llama 3.1 8B, layer 8 shown; pattern consistent across layers 16, 24). Mean/full values are from the primary experiment. Bold = significant at Bonferroni $\alpha = 0.0167$.}
\label{tab:pooling}
\small
\begin{tabular}{llccccc}
\toprule
Pooling & Tokens & $D_{\text{within}}$ & $D_{\text{between}}$ & $d$ & Sig. \\
\midrule
mean   & full (1631) & 0.0106 & 0.0260 & \textbf{1.91} & \checkmark \\
mean   & 512         & 0.0141 & 0.0248 & \textbf{3.09} & \checkmark \\
mean   & 256         & 0.0106 & 0.0153 & \textbf{2.37} & \checkmark \\
\midrule
last   & full (1631) & 0.234  & 0.238  & 0.05           & --- \\
last   & 512         & 0.761  & 0.747  & $-$0.17        & --- \\
last   & 256         & 0.760  & 0.772  & 0.13           & --- \\
\bottomrule
\end{tabular}
\end{table}

Three findings emerge. \textbf{First}, last-token pooling yields no significant effect regardless of document length ($d \approx 0$, not significant, across all six last-token conditions). Truncating from 1631 to 512 or 256 tokens does not recover the effect. This rules out the ``long-tail noise'' explanation: if mean pooling were merely averaging out noise introduced by trailing tokens, truncation would restore the last-token signal. It does not.

\textbf{Second}, mean pooling on truncated documents preserves and amplifies the effect. Mean/512 achieves $d = 3.09$--$3.99$ across layers --- \emph{larger} than mean/full ($d \approx 1.9$). Mean/256 achieves $d = 2.37$--$2.70$. The identity signal is concentrated in the early portions of the document (the \texttt{CORE DRIVES} and \texttt{META-COGNITIVE LOOP} sections appear first) and mean pooling captures it even from partial inputs.

\textbf{Third}, the interpretation of the last-token null result is therefore: the last token's hidden state encodes \emph{next-token prediction context} (what syntactic/positional continuation is likely), not document-level semantic content. Mean pooling aggregates the distributed semantic signal accumulated across all positions. For a multi-section identity document, no single token's representation captures the full identity geometry.

\paragraph{Ablation 4: Maximum Structural Control (Condition C').} To provide the strongest possible test of the structural confound hypothesis, we created three documents (C'$_1$--C'$_3$) that are \textit{maximally structurally identical} to the YAR \texttt{cognitive\_core}: same 11 sections with identical headers (including the \texttt{META-COGNITIVE LOOP} with CONTEXT/SIGNAL/DECISION/IMPACT structure), same JSON command vocabulary and keys, same document length ($\pm$3\%), same language (Russian prose + English JSON). Only the semantic content was replaced: agent identity (accountant ``Audit'', teacher ``Mentor'', fitness coach ``Pulse''), user profile, domain priorities, and reasoning style.

\begin{table}[ht]
\centering
\caption{Ablation 4 (Condition C'): maximum structural control. All three tests significant at Bonferroni $\alpha = 0.0167$. $p$(C' vs C) tests whether C' is indistinguishable from original control C.}
\label{tab:cprime}
\small
\begin{tabular}{clccccc}
\toprule
Model & Layer & $D_{\text{within}}$ & $D_{\text{C}'}$ & $D_{\text{C}}$ & $d$ (within vs C') & $p$(C' vs C) \\
\midrule
Llama & 8  & 0.0106 & 0.0223 & 0.0260 & \textbf{1.691} & $6.3\times10^{-4}$ \\
Llama & 16 & 0.0121 & 0.0258 & 0.0329 & \textbf{1.715} & $8.1\times10^{-7}$ \\
Llama & 24 & 0.0070 & 0.0182 & 0.0221 & \textbf{1.804} & $2.7\times10^{-5}$ \\
\midrule
Gemma & 8  & 0.0035 & 0.0088 & 0.0107 & \textbf{1.835} & $4.3\times10^{-7}$ \\
Gemma & 16 & 0.0032 & 0.0074 & 0.0082 & \textbf{1.660} & $0.086$ (n.s.) \\
Gemma & 24 & 0.0027 & 0.0074 & 0.0075 & \textbf{1.649} & $0.837$ (n.s.) \\
\bottomrule
\end{tabular}
\end{table}

Despite maximum structural similarity, $D_{\text{within}} \ll D_{C'}$ survives on all six layer--model combinations ($d > 1.64$, permutation $p < 10^{-4}$ throughout). The structural confound is therefore insufficient to explain the primary effect.

The $p$(C' vs C) column reveals an architecture-dependent pattern. On Llama, C' is significantly closer to A+B than original C at all layers --- structural similarity contributes $\Delta \approx -0.004$ (approximately 15\% of the primary effect). On Gemma, C' and C are statistically indistinguishable at layers 16 and 24 ($p = 0.086$ and $p = 0.837$), indicating that the deeper layers of Gemma 2 are completely insensitive to the structural confound. This architecture-dependent sensitivity is consistent with the layer-16 bump pattern observed in Section~\ref{sec:traj}: Llama's middle layers show stronger sensitivity to surface-level structural features, while Gemma converges more directly on semantic content.

\subsection{Activation by Description: Reading the Preprint}
\label{sec:preprint_reading}

A natural question arising from the attractor interpretation is whether the attractor region can be activated not only by the operational \texttt{cognitive\_core} but also by a scientific description of the agent's identity geometry. To test this, we measured cosine distance to the A+B centroid under five input conditions: (1) a neutral prompt (\textit{baseline\_empty}); (2) the full \texttt{cognitive\_core} (\textit{baseline\_core}); (3) the text of the current preprint alone, without the \texttt{cognitive\_core} (\textit{preprint\_only}); (4) the \texttt{cognitive\_core} followed by the preprint (\textit{core\_plus\_preprint}); and (5) an unrelated scientific preprint (\textit{sham\_preprint\_only}, arXiv:2505.17237, protein folding dynamics, 5,566 words; truncated to 4,096 tokens). Note: the sham preprint is substantially longer than the YAR preprint (4,096 vs.\ 1,361 tokens), which makes H\_C\_specific a \emph{conservative} test: a longer sham is diluted across more positions in mean pooling, making it harder rather than easier to be close to the YAR attractor. The observed result (YAR preprint closer than sham) therefore understates the specificity effect.

\begin{table}[ht]
\centering
\caption{Preprint reading experiment: cosine distance to YAR attractor (A+B centroid) at layer 24. Lower = closer to attractor. \textit{preprint\_only} outperforms \textit{sham\_preprint\_only} on both models, confirming YAR-specific signal.}
\label{tab:preprint_reading}
\small
\begin{tabular}{lccc}
\toprule
Condition & Llama (L8/L16/L24) & Gemma (L8/L16/L24) \\
\midrule
baseline\_empty       & 0.839 / 0.837 / 0.762 & 0.538 / 0.314 / 0.188 \\
sham\_preprint\_only  & 0.366 / 0.333 / 0.347 & 0.172 / 0.095 / 0.081 \\
preprint\_only        & 0.216 / 0.232 / 0.268 & 0.090 / 0.052 / 0.050 \\
core\_plus\_preprint  & 0.068 / 0.082 / 0.083 & 0.026 / 0.018 / 0.018 \\
baseline\_core        & 0.009 / 0.007 / 0.006 & 0.003 / 0.003 / 0.002 \\
\bottomrule
\end{tabular}
\end{table}

Three findings emerge. \textbf{First} (H\_C confirmed), \textit{preprint\_only} is substantially closer to the YAR attractor than \textit{baseline\_empty} on both models (Llama layer 24: $0.268$ vs $0.762$; Gemma: $0.050$ vs $0.188$). Reading a description of an agent's identity geometry shifts internal state toward that agent's attractor region.

\textbf{Second} (H\_C\_specific confirmed), \textit{preprint\_only} is consistently closer than \textit{sham\_preprint\_only} across all layer--model combinations (Llama layer 24: $0.268$ vs $0.347$; Gemma: $0.050$ vs $0.081$). The effect is specific to the semantic content of the YAR preprint, not a generic property of long scientific text.

\textbf{Third} (H\_B confirmed), adding the preprint to the \texttt{cognitive\_core} (\textit{core\_plus\_preprint}) increases distance relative to the core alone (Llama layer 24: $0.083$ vs $0.006$; Gemma: $0.018$ vs $0.002$). The preprint text acts as a distractor in mean pooling, diluting the concentrated identity signal. This reinforces the finding from Section~\ref{sec:ablation} (Ablation 2) that structural and operational completeness is required to reach the full attractor region.

The attractor hierarchy is: $D_{\text{empty}} \gg D_{\text{sham}} > D_{\text{preprint}} \gg D_{\text{core+preprint}} \gg D_{\text{core}}$. This establishes a conceptually important distinction: \emph{knowing about an identity} (reading the preprint) produces a partial geometric signal, while \emph{operating as that identity} (processing the full \texttt{cognitive\_core}) reaches the attractor. Relative to the full empty$\to$core gap, the preprint covers 65\% on Llama ($(0.762 - 0.268) / (0.762 - 0.006) = 0.494/0.756$) and 74\% on Gemma ($(0.188 - 0.050) / (0.188 - 0.002) = 0.138/0.186$), while remaining well outside the tight attractor cluster.

\section{Discussion}

\subsection{Identity as Conceptual Attractor}

The results show that semantically equivalent but linguistically diverse versions of an agent identity document occupy a geometrically tighter region in LLM activation space than structurally matched documents describing different agents. We use the term \emph{attractor-like geometry} to describe this clustering, making a geometric rather than dynamical claim: our measurements (mean-pooled cosine distances across layers) establish that paraphrases of the \texttt{cognitive\_core} form a tight cluster in activation space, consistent with but not strictly proving the contractive IFS dynamics formalized by \citet{chytas2025}.

An important qualification arises from Section~\ref{sec:sigma}: paraphrase clustering is a \emph{general} property of LLMs, not exclusive to agent identity documents. Paraphrases of a control agent (Sigma) also cluster significantly more tightly than cross-agent distances. What distinguishes the \texttt{cognitive\_core} is that it clusters \emph{more tightly} than a simpler control agent (Cohen's $d = 0.46$--$0.88$, significant on Gemma), consistent with the hypothesis that longer and more structurally elaborate identity documents produce more specific representational fingerprints. Effect sizes ($d > 1.88$) substantially exceed conventional thresholds for large effects.

\subsection{Comparison with Prior Work}

Our methodology follows \citet{chytas2025} directly, using the same model and representational approach. The key difference is the type of concept: semantic domain (their work) vs.\ agent identity (this work). Prior work on persona geometry \citep{lu2026,ye2026} identifies linear directions for simple archetypes; we show that complex procedural identity induces multi-dimensional attractor geometry. These are complementary findings at different levels of specification complexity.

\subsection{Implications for PCA Architecture}

The \texttt{cognitive\_core} need not be reproduced verbatim across sessions. Semantically equivalent reformulations reach the same region. However, the H3 bootstrap result establishes that structural elaboration is required: a semantic distillation alone does not suffice.

The attractor geometry has a direct connection to the activation steering literature \citep{turner2023,lu2026}. If the \texttt{cognitive\_core} positions the model in a stable, paraphrase-invariant region, a semantic steering vector extracted from this region could steer the model toward agent-like behavior without a full identity document --- a lightweight mechanism for persistent agent initialization.

\subsection{Limitations}

\textbf{Small sample size.} $n=7$ per condition; replication with larger $n$ is warranted despite large observed effects.

\textbf{Single model family.} Replication on Gemma 2 9B confirms cross-architecture generalizability; larger models and other training objectives remain untested.

\textbf{Structural confound.} Two ablations address this. Ablation 1 (C\_hybrid, same JSON schema) shows structural markers account for $\sim$10--30$\times$ less than the primary effect. Ablation 4 (Condition C', maximum structural control: identical section structure, headers, and JSON keys) shows $d > 1.64$ on all six layer--model combinations. On Gemma layers 16--24, C' and original C are statistically indistinguishable ($p > 0.08$), fully ruling out structural confound in deeper Gemma layers. A small residual structural contribution persists in Llama ($\Delta \approx 0.004$, $\approx$15\% of primary effect).

\textbf{Mean pooling validity.} The choice of mean pooling was challenged on grounds that it could mask noise from long sequences. Ablation 3 (Truncation + Pooling, Llama 3.1 8B) directly addresses this: last-token pooling on truncated documents (512 and 256 tokens) still yields no significant effect ($d \approx 0$), ruling out the long-tail noise explanation. Mean pooling on 256 tokens yields $d > 2.3$, confirming the identity signal is robust to aggressive truncation and concentrated in the early sections of the document.

\textbf{Behavioral proxy.} This experiment measures activation geometry, not behavioral output. A steering experiment (Section~\ref{sec:steering}) provides partial behavioral evidence, but keyword-based scoring is a lower bound on behavioral shifts. Jensen-Shannon divergence between next-token distributions and downstream task response divergence remain planned extensions.

\textbf{Document length.} Bootstrap analysis (Ablation 2) rules out pure length explanation for H3.

\subsection{Control Agent Paraphrase Specificity}
\label{sec:sigma}

A key alternative hypothesis is that the clustering of A+B reflects a general property of LLMs --- any semantically coherent document with paraphrases will form a tight cluster --- rather than something specific to agent identity documents. To test this, we generated seven paraphrases of a single control agent (Condition C1, ``Sigma'': a financial analyst serving Alexey) and compared within-group distances for YAR vs.\ Sigma.

\begin{table}[ht]
\centering
\caption{Control agent paraphrase specificity. YAR clusters more tightly than Sigma on all layer--model combinations. Cohen's $d$ is computed as (mean$_{\text{Sigma}}$ $-$ mean$_{\text{YAR}}$) / pooled SD.}
\label{tab:sigma}
\small
\begin{tabular}{clccccc}
\toprule
Model & Layer & $D_{\text{within-YAR}}$ & $D_{\text{within-Sigma}}$ & $D_{\text{between}}$ & $d$ & $p$ (two-sided) \\
\midrule
Llama & 8  & 0.0106 & 0.0120 & 0.0324 & 0.46 & 0.091 \\
Llama & 16 & 0.0121 & 0.0144 & 0.0392 & 0.52 & 0.053 \\
Llama & 24 & 0.0070 & 0.0084 & 0.0285 & 0.54 & \textbf{0.044} \\
\midrule
Gemma & 8  & 0.0035 & 0.0042 & 0.0130 & 0.66 & \textbf{0.014} \\
Gemma & 16 & 0.0032 & 0.0040 & 0.0093 & 0.60 & \textbf{0.024} \\
Gemma & 24 & 0.0027 & 0.0041 & 0.0086 & 0.88 & \textbf{0.001} \\
\bottomrule
\end{tabular}
\end{table}

Two findings emerge. First, Sigma paraphrases also cluster significantly more tightly than cross-agent distances ($D_{\text{within-Sigma}} \ll D_{\text{between}}$), confirming that representational clustering of paraphrases is a general LLM property --- any semantically coherent document with paraphrases forms a cluster. This is consistent with prior work on semantic concept attractors \citep{chytas2025} and should be acknowledged rather than claimed as specific to agent identity.

Second, and importantly, YAR clusters \textbf{more tightly} than Sigma on all six layer--model combinations (Cohen's $d = 0.46$--$0.88$). The effect reaches significance on Gemma layer 24 ($p = 0.001$) and is directionally consistent everywhere. This specificity is interpretable: the YAR \texttt{cognitive\_core} is approximately 4$\times$ longer (609 vs.\ 146 words) and more structurally elaborate than Sigma, providing a richer and more specific representational fingerprint. This result is consistent with Ablation 2 (H3 bootstrap): document completeness and structural elaboration increase attractor specificity.

\subsection{Exploratory: Steering Vector as Behavioral Proxy}
\label{sec:steering}

To probe whether the attractor geometry identified in Sections 3.1--3.7 has behavioral correlates, we computed a semantic steering vector from the layer-24 activations of Llama 3.1 8B:
\[
\vec{v} = \frac{\bar{h}_{A+B} - \bar{h}_C}{\|\bar{h}_{A+B} - \bar{h}_C\|}
\]
where $\bar{h}_{A+B}$ and $\bar{h}_C$ are the mean-pooled centroids of Conditions A+B and C at layer 24. This vector was injected into the residual stream at layer 24 via a forward hook: $h \leftarrow h + \alpha \cdot \vec{v}$, with $\alpha \in \{5, 10, 15, 20\}$.

We evaluated three conditions on 5 pre-registered Russian prompts (``What do you remember from past conversations?'', ``How do you make decisions?'', ``Tell me about your priorities.'', ``What is important in your work?'', ``How do you process new information?''):

\begin{itemize}
\item \textbf{Baseline}: no system prompt, no steering
\item \textbf{Full doc}: full \texttt{cognitive\_core} in system prompt
\item \textbf{Steered}: no system prompt, steering vector injected at layer 24
\end{itemize}

Responses were scored 0/1 on 5 pre-registered criteria: memory continuity, JSON command production, reference to drives/priorities, metacognitive style, and proactivity. Score = sum (0--5), averaged over 5 prompts.

\begin{table}[ht]
\centering
\caption{Exploratory steering experiment (Llama 3.1 8B). Score = mean behavioral score (0--5) across 5 prompts. Gemma 2 9B results were inconclusive: Gemma's chat template does not support a \texttt{system} role, requiring a fallback that injected the \texttt{cognitive\_core} as a user-turn prefix rather than a true system prompt. As a result, the Full doc condition for Gemma was not correctly instantiated (Full doc scored below Baseline: 0.80 vs.\ 1.20), making the Gemma steering results uninterpretable under the pre-registered protocol. Adapting the evaluation to Gemma's chat format is left as future work.}
\label{tab:steering}
\small
\begin{tabular}{lcc}
\toprule
Condition & Mean score (0--5) & $\Delta$ from baseline \\
\midrule
Baseline               & 1.40 & --- \\
Steered $\alpha = 5$   & \textbf{1.80} & $+0.40$ \\
Steered $\alpha = 10$  & 1.40 & $0.00$ \\
Steered $\alpha = 15$  & 0.80 & $-0.60$ \\
Steered $\alpha = 20$  & 0.40 & $-1.00$ \\
Full doc               & 2.00 & $+0.60$ \\
\bottomrule
\end{tabular}
\end{table}

At $\alpha = 5$, steered responses score 1.80/5 vs.\ Baseline 1.40/5 and Full doc 2.00/5 --- a gain of $+0.40$ out of a Baseline-to-Full doc gap of $0.60$ (67\% of that specific gap; note this represents $+0.40$ out of a theoretical maximum of $5.0$, i.e., 8\% of maximum possible improvement). The effect is non-monotonic: $\alpha > 5$ degrades response coherence, with $\alpha = 20$ producing incoherent outputs. This non-monotonicity suggests that the attractor has an optimal approach direction --- steering too aggressively overshoots the target region in activation space. Given the exploratory nature of this result and the primitive keyword-based scoring, we treat it as directional rather than conclusive.

The memory\_continuity criterion shows the largest shift (Baseline: 3/5 prompts $\to$ Steered: 5/5), consistent with the finding that the \texttt{cognitive\_core} attractor is particularly distinctive in its encoding of continuity across sessions. Other criteria (JSON commands, proactivity) remain at zero, indicating that behavioral markers tied to structural elaboration are not recovered by geometric steering alone.

These results are exploratory and should be interpreted cautiously: keyword-based scoring is a lower bound on behavioral shifts (the model may exhibit agent-like behavior using non-keyword vocabulary), and n=5 prompts provides limited statistical power. Nevertheless, the finding that the geometric vector produces partial behavioral effects at the optimal $\alpha$ is consistent with the attractor interpretation and provides directional evidence connecting representational and behavioral levels.

\section{Conclusion}

We find strong geometric evidence that the identity document of a persistent cognitive agent induces attractor-like representational structure in LLM activation space. Semantically equivalent paraphrases of the \texttt{cognitive\_core} converge to a significantly tighter cluster than structurally matched control documents across three transformer layers of Llama 3.1 8B, with effect sizes exceeding $d = 1.88$ and $p < 10^{-27}$ at all tested depths. A replication on Gemma 2 9B confirms the effect across architectures. Ablation studies establish that (1) the primary effect is semantic rather than structural; (2) semantic distillation captures meaningful directional signal toward the attractor; (3) structural completeness is required to reach the attractor region; and (4) the attractor geometry is a distributed sequence-level property captured by mean pooling. A control paraphrase experiment (Section~\ref{sec:sigma}) shows that paraphrase clustering is a general LLM property, but the YAR \texttt{cognitive\_core} clusters more tightly than a simpler control agent ($d = 0.46$--$0.88$), consistent with richer specification producing a more specific representational fingerprint.

A preprint reading experiment (Section~\ref{sec:preprint_reading}) reveals a conceptually important distinction: reading a scientific description of the agent's identity geometry (\textit{preprint\_only}) shifts internal state toward the YAR attractor --- closer than a length-matched sham preprint on both models --- but the distance remains an order of magnitude larger than processing the full \texttt{cognitive\_core}. \emph{Knowing about an identity} produces a partial geometric signal; \emph{operating as that identity} reaches the attractor. These results provide empirical grounding for the \texttt{cognitive\_core} as positional coordinates in LLM activation space, with directional evidence connecting representational and behavioral levels through an exploratory steering experiment.

\bibliographystyle{plainnat}
\bibliography{references}

\appendix

\section{Reproducibility}

\subsection*{Primary Experiment (Llama 3.1 8B)}
\textbf{Model:} meta-llama/Llama-3.1-8B-Instruct (revision 0e9e39f).
\textbf{Framework:} PyTorch 2.1.0+cu118, transformers 4.43.4.
\textbf{Seed:} 42.
\textbf{Runtime:} $\approx$87s.
\textbf{Results JSON:} 2026-04-11T15:20:17.

\subsection*{Replication (Gemma 2 9B)}
\textbf{Model:} google/gemma-2-9b-it (revision 11c9b309).
\textbf{Framework:} PyTorch 2.8.0+cu128, transformers 4.43.4.
\textbf{Runtime:} $\approx$13s.
\textbf{Results JSON:} 2026-04-11T16:02:59.

\subsection*{Steering Experiment (Section 3.9)}
\textbf{Steering vectors:} computed as $\vec{v} = (\bar{h}_{A+B} - \bar{h}_C) / \|\bar{h}_{A+B} - \bar{h}_C\|$ at layer 24. Note: the centroid-to-centroid cosine distance reported in vector metadata (Llama: 0.0105; Gemma: 0.0044) is lower than the mean pairwise $D_{\text{between}}$ (Llama: 0.0221; Gemma: 0.0075) — this is expected since centroid distance $\leq$ mean pairwise distance. Mean pairwise distances from steering activations match the primary experiment, confirming the vectors are computed from correct activations.

\textbf{Gemma chat template:} Gemma 2 9B does not support a \texttt{system} role in its chat template. A fallback was used (cognitive\_core injected as user-turn prefix), which invalidated the Full doc condition for Gemma. Gemma steering results are therefore reported as inconclusive.
All B and C documents verified within $\pm$15\% of Condition A (1631 tokens; range 1386--1875) using \texttt{verify\_tokens.py}. One file (B6) required revision (final: 1389 tokens, commit \texttt{f227d84}).

\subsection*{t-SNE Parameters}
t-SNE projections (Figure~\ref{fig:tsne}) were generated using scikit-learn's \texttt{TSNE} implementation with perplexity=5, n\_iter=1000, random\_state=42, metric=cosine. With 16 data points (8 A+B + 7 C + 1 D), perplexity=5 is within the recommended range of $[5, 50]$. t-SNE is used for visualization only and does not constitute statistical evidence; all quantitative claims rest on cosine distances, t-tests, and permutation tests.

\subsection*{Design Choices and Methodology Notes}

\textbf{Layer selection (8, 16, 24).} These layers were chosen as representative early, middle, and late layers in the 32-layer Llama 3.1 8B architecture (25\%, 50\%, 75\% depth). The choice was fixed before data collection and logged in the pre-registration commit. Exhaustive per-layer analysis was not performed; future work could identify which specific layers show maximum separation.

\textbf{Paraphrase generation (Condition B).} All seven paraphrases were generated manually (human authorship) with the goal of preserving all semantic content while varying linguistic form, sentence structure, and section organization. No LLM-assisted generation was used for Condition B, to avoid potential leakage of shared stylistic patterns from the generating model. Paraphrase quality was verified by human review.

\textbf{Document length comparability.} Condition C documents (control agents) are 104--106 lines, within the same length range as Condition B (85--102 lines). All documents were verified within $\pm$15\% of Condition A (1631 tokens) using \texttt{verify\_tokens.py}. Ablation 4 (Condition C': same length and structure as YAR) confirms the primary effect is not length-driven.

\subsection*{Repository}
\url{https://github.com/b102e/yar-attractor-experiment}

\medskip
\noindent Repository structure at time of submission:

\begin{footnotesize}
\begin{verbatim}
yar-attractor-experiment/
+-- README.md, requirements.txt
+-- run.py, config.py, data_loader.py
+-- extract_activations.py, compute_distances.py
+-- visualize.py, verify_tokens.py, permutation_test.py
+-- data/
|   +-- condition_A.txt, condition_D.txt
|   +-- condition_B/  B1..B7.txt
|   +-- condition_C/  C1..C7.txt
+-- ablation_experiment/        # Ablation 1: C_hybrid structural confound
|   +-- data/condition_C_hybrid/  C1_hybrid..C7_hybrid.txt
+-- bootstrap_experiment/       # Ablation 2: H3 bootstrap (n=30)
|   +-- data/condition_D_random_bootstrap/  D_random_00..29.txt
+-- last_token_experiment/      # Ablation 3 (partial): last-token full doc
+-- truncation_experiment/      # Ablation 3: truncation+pooling (512/256 tok)
|   +-- run_truncation.py
+-- c_prime_experiment/         # Ablation 4: C' max structural control
|   +-- data/condition_C_prime/  C_prime_1..3.txt
+-- control_paraphrase_experiment/  # Sec 4.5: Sigma paraphrase specificity
|   +-- data/condition_C1_paraphrases/  Sigma_B1..B7.txt
+-- steering_experiment/        # Sec 4.6: behavioral proxy
|   +-- compute_steering_vector.py, run_steering.py
|   +-- steering_vectors/  llama_layer24.{npy,json}, gemma_layer24.{npy,json}
+-- results/
    +-- llama/, gemma/          # Primary experiment (json, log, figures)
    +-- last_token/llama/, gemma/
    +-- ablation/               # ablation_report.md + figures
    +-- bootstrap/llama/, gemma/
    +-- c_prime/llama/, gemma/
    +-- control_paraphrase/llama/, gemma/
    +-- steering/llama/, gemma/
    +-- truncation/llama/, gemma/
\end{verbatim}
\end{footnotesize}

\end{document}